\documentclass[runningheads]{llncs}
\usepackage{graphicx}

\usepackage{tikz}
\usepackage{comment}
\usepackage{amsmath,amssymb} 
\usepackage{color}
\usepackage{subfigure}
\usepackage{booktabs}
\usepackage{multirow}
\usepackage[algo2e]{algorithm2e}
\usepackage[noend]{algpseudocode}
\usepackage{algorithmicx,algorithm}

\usepackage[accsupp]{axessibility}  

\begin{document}
\pagestyle{headings}
\mainmatter
\def\ECCVSubNumber{2316}  

\title{S4OD: Semi-Supervised learning for Single-Stage Object Detection} 

\titlerunning{S4OD: Semi-Supervised learning for Single-Stage Object Detection.}
%
\author{Yueming Zhang\inst{1,2} \and
Xingxu Yao\inst{2} \and
Chao Liu\inst{2} \and
Feng Chen\inst{2} \and
Xiaolin Song\inst{2} \and
Tengfei Xing\inst{2} \and
Runbo Hu\inst{2} \and
Hua Chai\inst{2} \and
Pengfei Xu\inst{2} \and
Guoshan Zhang\inst{1}}
\authorrunning{Yueming Zhang et al.}
%
\institute{Tianjin University \\ \email{\{seife, zhanggs\}@tju.edu.cn} \and Didi Chuxing}
\maketitle
\begin{abstract}
Single-stage detectors suffer from extreme foreground-background class imbalance, while two-stage detectors do not. 
Therefore, in semi-supervised object detection, two-stage detectors can deliver remarkable performance by only selecting high-quality pseudo labels based on classification scores.
However, directly applying this strategy to single-stage detectors would aggravate the class imbalance with fewer positive samples. 
Thus, single-stage detectors have to consider both quality and quantity of pseudo labels simultaneously. 
In this paper, we design a dynamic self-adaptive threshold (DSAT) strategy in classification branch, which can automatically select pseudo labels to achieve an optimal trade-off between quality and quantity. 
Besides, to assess the regression quality of pseudo labels in single-stage detectors, we propose a module to compute the regression uncertainty of boxes based on Non-Maximum Suppression. 
By leveraging only 10\% labeled data from COCO, our method achieves 35.0\% AP on anchor-free detector (FCOS) and 32.9\% on anchor-based detector (RetinaNet).

\end{abstract}

\section{Introduction}
The performance of deep learning networks benefit from tremendous data \cite{dosovitskiy2020image}. However, it's expensive and intractable to obtain a large-scale human-labeled dataset for a new application scenario. This issue has promoted semi-supervised learning to become the research topic that receives substantial interests in various fields \cite{chen2021semi,assran2021semi,jiang2019semi,zhai2019s4l,xu2021end}. For object detection tasks with significant application value yet high annotation cost, it is of great significance to research on semi-supervised object detection.

\begin{figure}[tb]
    \centering
    \subfigure[]{
    \begin{minipage}[t]{0.45\linewidth}
        \centering
        \includegraphics[width=\linewidth]{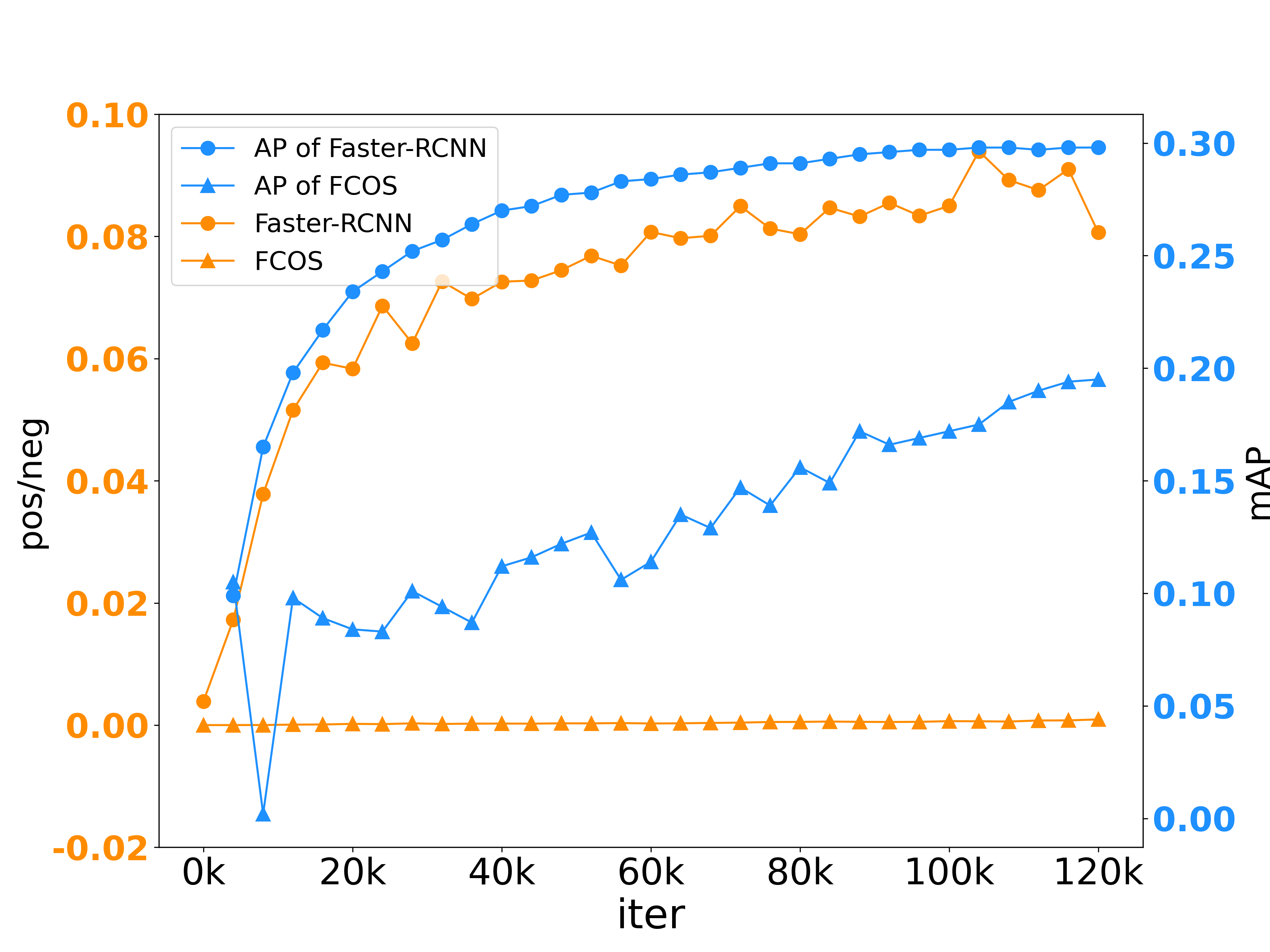}
        \label{ratio_figure}
    \end{minipage}
    }
    \subfigure[]{
    \begin{minipage}[t]{0.45\linewidth}
        \centering
        \includegraphics[width=\linewidth]{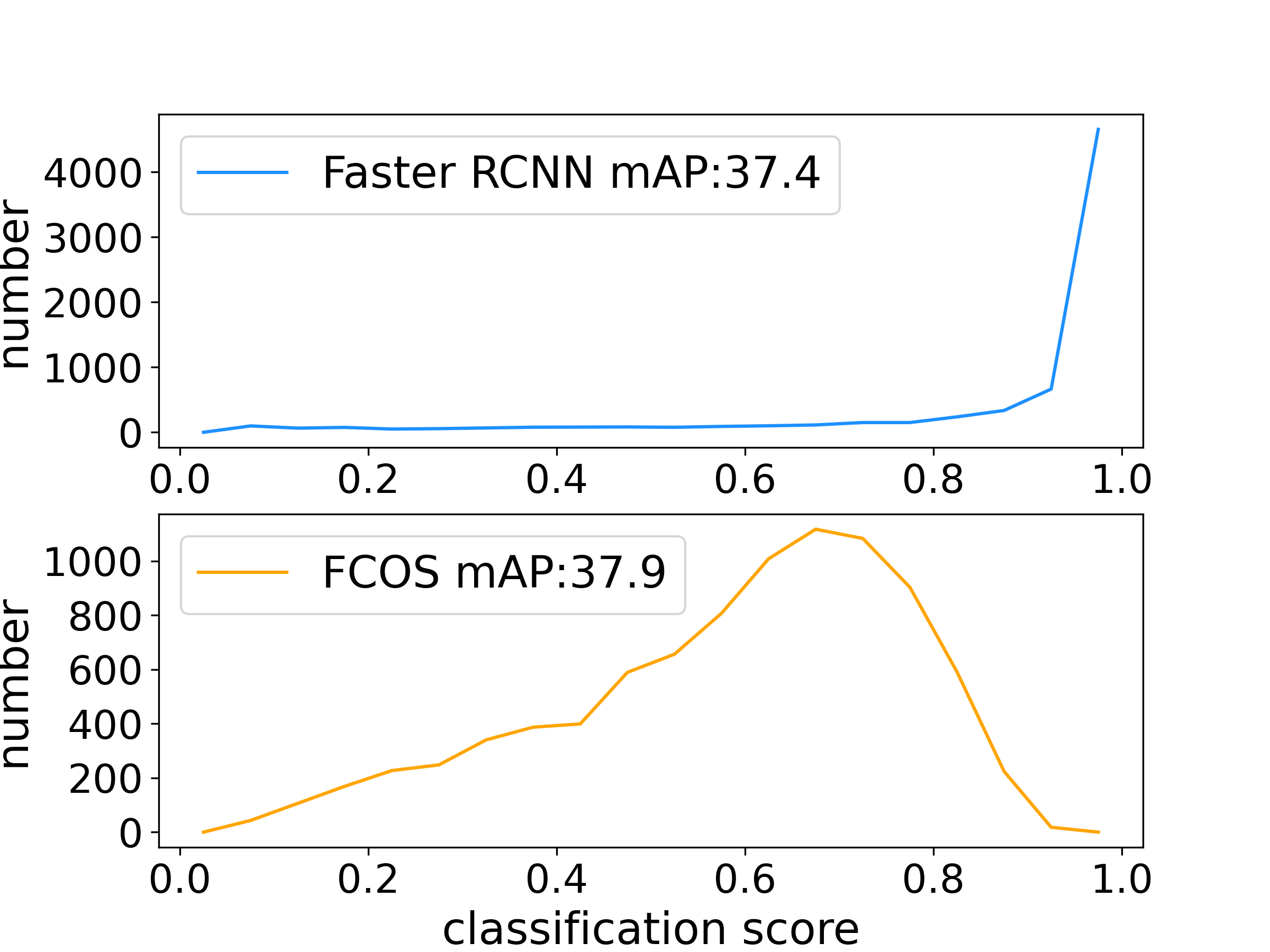}
        \label{distribution_difference}
    \end{minipage}
    }
    \caption{\textbf{(a):} Compared with Faster-RCNN, FCOS suffers from a serious sampling imbalance (\textcolor[RGB]{240,145,53}{orange triangle}) for pseudo labels which further leads to poor performance (\textcolor[RGB]{66,144,247}{blue triangle}). \textcolor[RGB]{240,145,53}{Orange lines} are the ratio of positives and negatives in FCOS and that in Faster-RCNN on pseudo labels, and the \textcolor[RGB]{66,144,247}{blue lines} are the mAP curves on COCO-val correspondingly.  
    \textbf{(b):} The distribution of classification score on COCO-val, only predictions with IoU\textgreater 0.9 are counted.}
    \label{nouse}
\end{figure}

Most previous studies focus on the two-stage detectors \cite{jeong2019consistency,liu2021unbiased,sohn2020simple,zhou2021instant,yang2021interactive,xu2021end,wang2021combating}. \cite{xu2021end,liu2021unbiased} state that their methods could be applied to single-stage object detectors in theory, but few empirical results can systematically support this. As proved in \cite{lin2017focal}, the major difference between single- and two-stage detectors is the class balance difference. Specifically, in two-stage detectors, region proposal network (RPN) filters out most of background samples, and a fixed ratio is adopted in the next stage to maintain a sampling balance between foregrounds and backgrounds. However, single-stage detectors directly make dense predictions covering all space locations in each scale. Only a small fraction of predictions are assigned as positive samples. This leads to a severe imbalance of positive and negative samples. In semi-supervised learning, the improper pseudo labels selection strategy would aggravate this imbalance. For example, as shown in Fig.\ref{ratio_figure}, directly using the same setting from two-stage detectors in single-stage detectors would drastically sharpen the number of positive samples, pushing the ratio of positives and negatives towards zero. Such imbalance forces the detectors to make predictions with an inductive bias on the background class, and therefore limits the performance of semi-supervised single-stage detectors.

To alleviate the sampling imbalance of single-stage detectors in semi-supervised learning, we focus on the strategy of selecting pseudo labels. Conventional methods \cite{xu2021end,liu2021unbiased} use an ablation study to determine the strategy of pseudo label filtering, which is complicated and unexplainable. Most importantly, this strategy lowers the practical value of these semi-supervised learning methods. For example, \cite{xu2021end} and \cite{liu2021unbiased} both use Faster-RCNN \cite{ren2015faster}, while the most proper thresholds obtained from the ablation study are not identical (0.9 and 0.7). This case motivates us to find a more universal pseudo label selecting strategy. To this end, we first compare the classification distribution of final predictions between the single- and two-stage detectors. As shown in Fig.\ref{distribution_difference}, the classification score of single-stage detectors is significantly lower than that of two-stage detectors. This gap indicates that transferring the selective strategy of two-stage detectors to the single-stage detectors would reduce the number of pseudo labels and further lead to a more severe sampling imbalance, although the quality of pseudo labels is still guaranteed. In contrast, choosing a loose strategy would increase the number of pseudo labels and alleviate this imbalance, but more noises are brought and the quality of pseudo labels cannot be promised. Therefore, inspired by this insight, we propose a dynamic self-adaptive threshold (DSAT) which can automatically determine the selective strategy considering both quality and quantity of pseudo labels. More importantly, this strategy can dynamically select pseudo labels which are aligned with the model performance during the training process.

As another part of object detection, the regression quality of pseudo labels is of equal importance as the classification score. However, it is hard to evaluate the regression quality in continuous space. To avoid the impact caused by the low-quality pseudo labels, Liu et al. \cite{liu2021unbiased} abandon the regression loss when computing the unsupervised loss. Xu et al. \cite{xu2021end} find that the regression results are related to the uncertainty which is computed by the BoxJitter module. However, the BoxJitter module relies on the Region-CNN (RCNN) to process the jittered proposals, which means that it is infeasible for single-stage detectors. To address this issue, we define the standard deviation of the redundant bounding boxes in non-maximum suppression (NMS) as the corresponding box uncertainty and choose the pseudo labels with lower uncertainty as the regression target. The proposed module is named NMS-UNC as it is integrated with the original NMS module.

In summary, we propose a semi-supervised learning framework for single-stage detectors (S4OD). Our contributions can be summarized as follows:

\begin{itemize}
    \item We reveal the importance of pseudo label selecting strategy in semi-supervised object detection and propose a DSAT strategy to dynamically determine the strategy which takes both quality and quantity of pseudo labels into consideration.
    \item To assess the quality of box regression in single-stage detectors, we develop a NMS-UNC module which computes the uncertainty of boxes based on the original NMS. 
    \item Experiments of using different percentages of labeled data on MS-COCO dataset show our method achieves a significant improvement (e.g. \textbf{+10.1\%} AP using 1\% labeled data) than fully-supervised learning. 
\end{itemize}

\section{Related Works}
\subsection{Object Detection}
In general, CNN-based object detectors can be split into two categories: proposal-based two-stage detectors \cite{ren2015faster,cai2018cascade,lin2017feature,he2017mask,chen2019hybrid} and single-stage detectors \cite{wang2021learn,ge2021yolox,redmon2018yolov3,liu2016ssd,tian2019fcos,lin2017focal}. Two-stage detectors rely on the sparse sampling of RPN to filter out most background samples and further refine the class balanced proposals. However, single-stage detectors use a single convolution network to directly make dense predictions, inevitably leading to the sampling imbalance. Single-stage detectors are more efficient while the two-stage detectors take the lead in terms of accuracy. Another paradigm is transformer-based method \cite{liu2021swin,liu2021swin2,zhu2020deformable}, these detectors achieve excellent performance relying on the strong feature extracting capacity of transformer. In this paper, we focus on two typical single-stage detectors: FCOS \cite{tian2019fcos} for anchor-free method and RetinaNet \cite{lin2017focal} for anchor-based method.

\subsection{Semi-Supervised Learning in Object Detection}
Applying semi-supervised learning in object detection is more challenging than that in classification \cite{pham2021meta,sohn2020fixmatch,xie2019unsupervised,wei2021crest,berthelot2019mixmatch}, because object detection combines both classification and regression. CSD \cite{jeong2019consistency} explicitly applies the consistency regularization on both labeled and unlabeled data for semi-supervised object detection. STAC \cite{sohn2020simple} is a semi-supervised learning framework based on self-training strategy and augmentation-driven consistency regularization. Instant-Teaching \cite{zhou2021instant} is designed to be a two-model co-rectify framework for increasing the accuracy of pseudo labels during the training process. Wang et al. \cite{wang2021data} divide unlabeled data according to the difficulty level and utilizes data from different levels in different training phases. ISD \cite{jeong2021interpolation} proposes a novel interpolation-based semi-supervised learning algorithm for object detection and conducts experiments on SSD \cite{liu2016ssd} network, but only anchor-based single-stage detector is considered. In this paper, we employ the teacher-student mutual learning framework in an end-to-end manner without redundant implementation, and conduct experiments on both anchor-free and anchor-based single-stage object detectors. Our model significantly improves the application potential of semi-supervised learning in single-stage object detector.

\section{Preliminary of Semi-Supervised Learning}
\subsection{Mutual Learning Framework}
Mainstream semi-supervised methods \cite{liu2021unbiased,tang2021humble,xu2021end} for object detection use the teacher-student mutual learning framework, consisting of a student model $\theta^S$ and a teacher model $\theta^T$. For labeled data $D^l$, the student model $\theta^S$ uses its ground truth to compute loss $L_{sup}$ for gradient descent:
\begin{equation}
    L_{sup} = \frac{1}{N_{pos}}\sum_{x,y}L_{cls}(p_{x,y}, c_{x,y}) + \frac{1}{N_{pos}}\sum_{i}^{N_{pos}}L_{reg}(t_{i}, b_{i})
    \label{sup_loss}
\end{equation}
where $N_{pos}$ is the number of positive samples. $p_{x,y}$ and $c_{x,y}$ refer to the predicted logit and ground-truth at location $(x,y)$. $t_{i}$ denotes the predicted regression result of the $i$-th positive sample, while $b_{i}$ is the corresponding target.

As for unlabeled data, the teacher model $\theta^T$ firstly inferences results on the weak-augmented unlabeled data $D^u_{weak}$. Then a classification score threshold $\sigma_{cls}$ is determined to filter out low-quality results. The reserved high-quality prediction is regarded as the ground truth (pseudo labels) for the strong-augmented unlabeled data $D^u_{strong}$. Afterwards, the student model computes unsupervised loss $L_{unsup}$ as Eq. (\ref{unsup_loss}) for unlabeled data and updates parameters $\theta^S$ through gradient descent. Finally, the teacher model $\theta^T$ is updated by the student model $\theta^S$ with exponential moving average (EMA) \cite{tarvainen2017mean}.

\begin{equation}
    L_{unsup} = \frac{1}{N_{pos}}\sum_{x,y}L_{cls}(p_{x,y}, c^*_{x,y}) + \frac{1}{N_{pos}}\sum_{i}^{N_{pos}}L_{reg}(t_{i}, b^*_{i})
    \label{unsup_loss}
\end{equation}
where $c^*$ and $b^*$ are the categories and boxes of positive samples. These positive samples are assigned from the reserved pseudo labels with higher classification scores than threshold $\sigma_{cls}$.

\subsection{Difference between Single- and Two-Stage Detectors}
\label{method_1}
Two-stage detectors include a RPN to generate proposals and a cascaded network to predict classification scores and refine regression results of each proposal, while single-stage object detectors directly generate dense outputs to predict the classification scores and regression results for each location. 

In order to have a better insight into the semi-supervised implementation difference between single- and two-stage detectors, we visualise the classification confidence distribution of single- and two-stage detectors in Fig.\ref{distribution_difference}. As shown in Fig.\ref{distribution_difference}, although the performances are comparable (37.4\% and 37.9\% mAP), there is a significant difference in classification score distribution between the single- and two-stage detectors. The classification confidence in the two-stage detectors (Faster-RCNN \cite{ren2015faster}) is generally higher than that in the single-stage (FCOS \cite{tian2019fcos}). Specifically, most classification scores of the proposals in Faster-RCNN locate in $[0.9, 1.0)$, while that in FCOS locate in the range of $[0.6, 0.8]$. For two-stage detectors, with the filtering and sampling capacity of RPN, the second stage could generate strong confidence scores on plenty of reserved high-quality proposals. Hence, two-stage detectors could benefit from choosing a more strict selecting strategy of pseudo labels. This strategy ensures the quality of pseudo labels without losing too much pseudo labels. 
However, single-stage detectors make predictions on the imbalanced samples, so the confidence scores are not as high as that in two-stage. This difference indicates that the selective strategy of pseudo labels in single-stage detectors must consider both the quality and quantity of them. A more strict selective strategy could ensure the quality of pseudo labels but the amount of them decreased greatly. This would exacerbate the inherent imbalance of positive and negative samples. The imbalanced sampling ratio leads negative samples dominating gradient and push the network to generate predictions with bias on background. As the teacher model is updated by student model, the biased representation would be accumulated in the mutual learning framework. This issue would affect the generalization ability of model and undermine the converging speed. Conversely, a tolerant selective strategy would guarantee the quantity of pseudo labels, but more noises are introduced in mutual learning frameworks. The accumulated error would also generate adverse effects on detectors. Based on the above analysis, we believe that the strategy of selecting pseudo labels is the dominant reason for limiting the performance in semi-supervised single-stage object detectors.

\begin{figure}[tb]
    \centering
    \includegraphics[width=\linewidth]{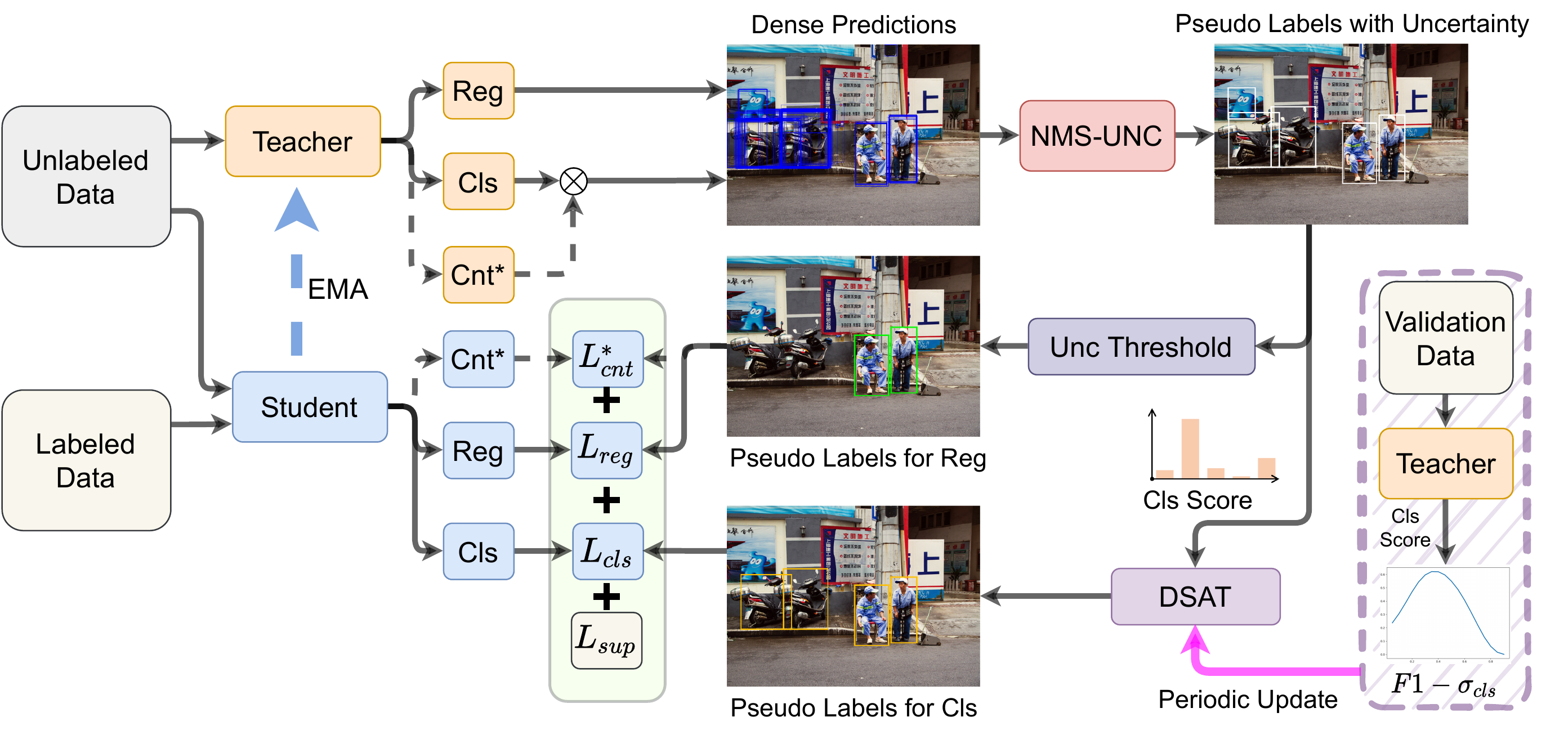}
    \caption{Pipeline of the proposed S4OD. \textbf{Reg} and \textbf{Cls} refer to the regression result and classification result. \textbf{Cnt*} denotes the centerness result for FCOS. \textbf{NMS-UNC} computes the uncertainty by Eq. (\ref{unc_com}) and outputs the sparse pseudo labels (White boxes). An \textbf{Unc threshold} is used to preserve high-quality regression boxes (Green boxes). \textbf{DSAT} dynamically adjusts the strategy according to the $F1-\sigma_{cls}$ distribution and select the boxes with high classification scores (Orange boxes).}
    \label{pipeline}
\end{figure}

\begin{figure}[tb]
    \centering
    \subfigure[]{
    \begin{minipage}[t]{0.45\linewidth}
        \centering
        \includegraphics[width=\linewidth]{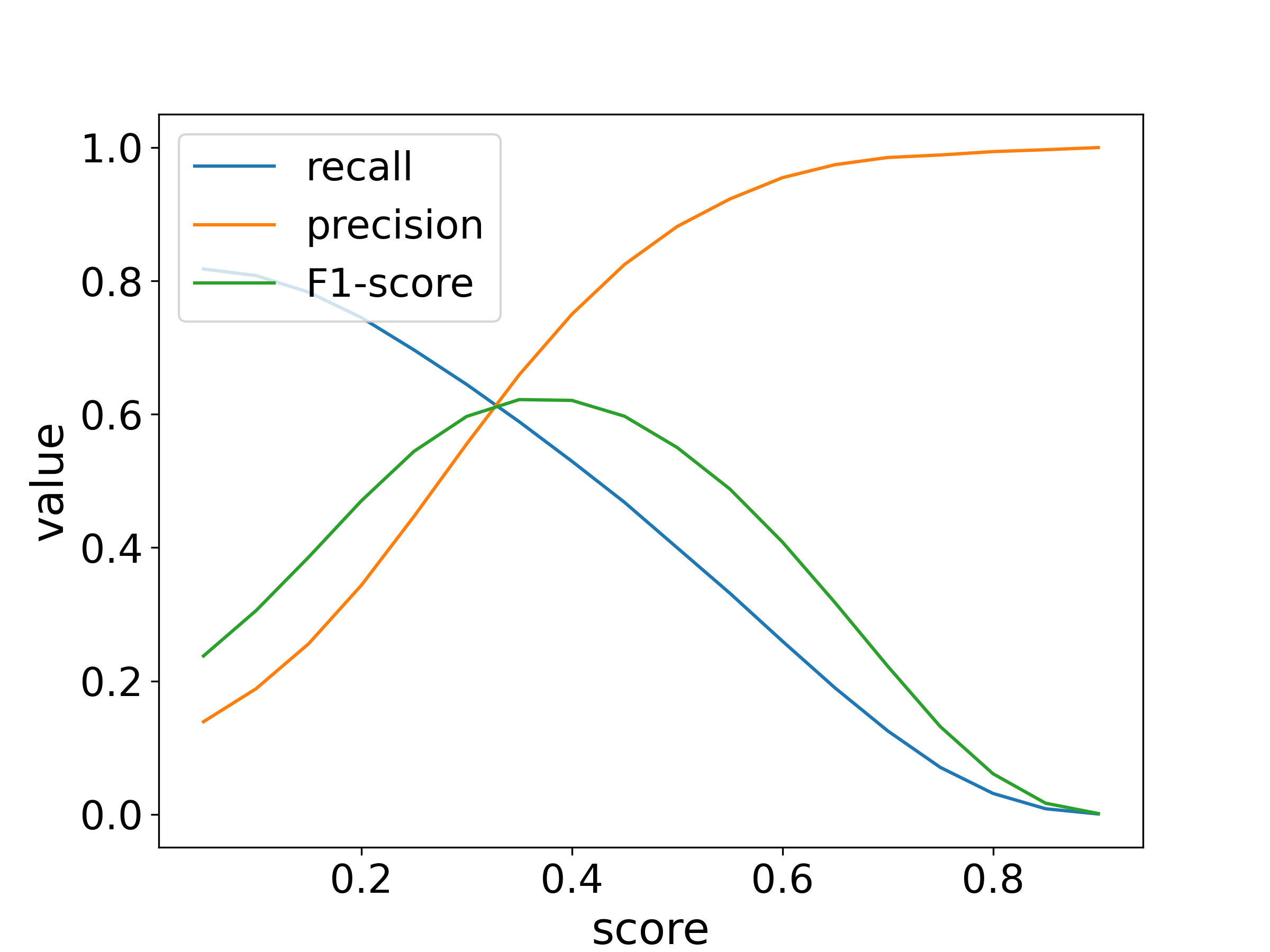}
        \label{epoch22}
    \end{minipage}
    }
    \subfigure[]{
    \begin{minipage}[t]{0.45\linewidth}
        \centering
        \includegraphics[width=\linewidth]{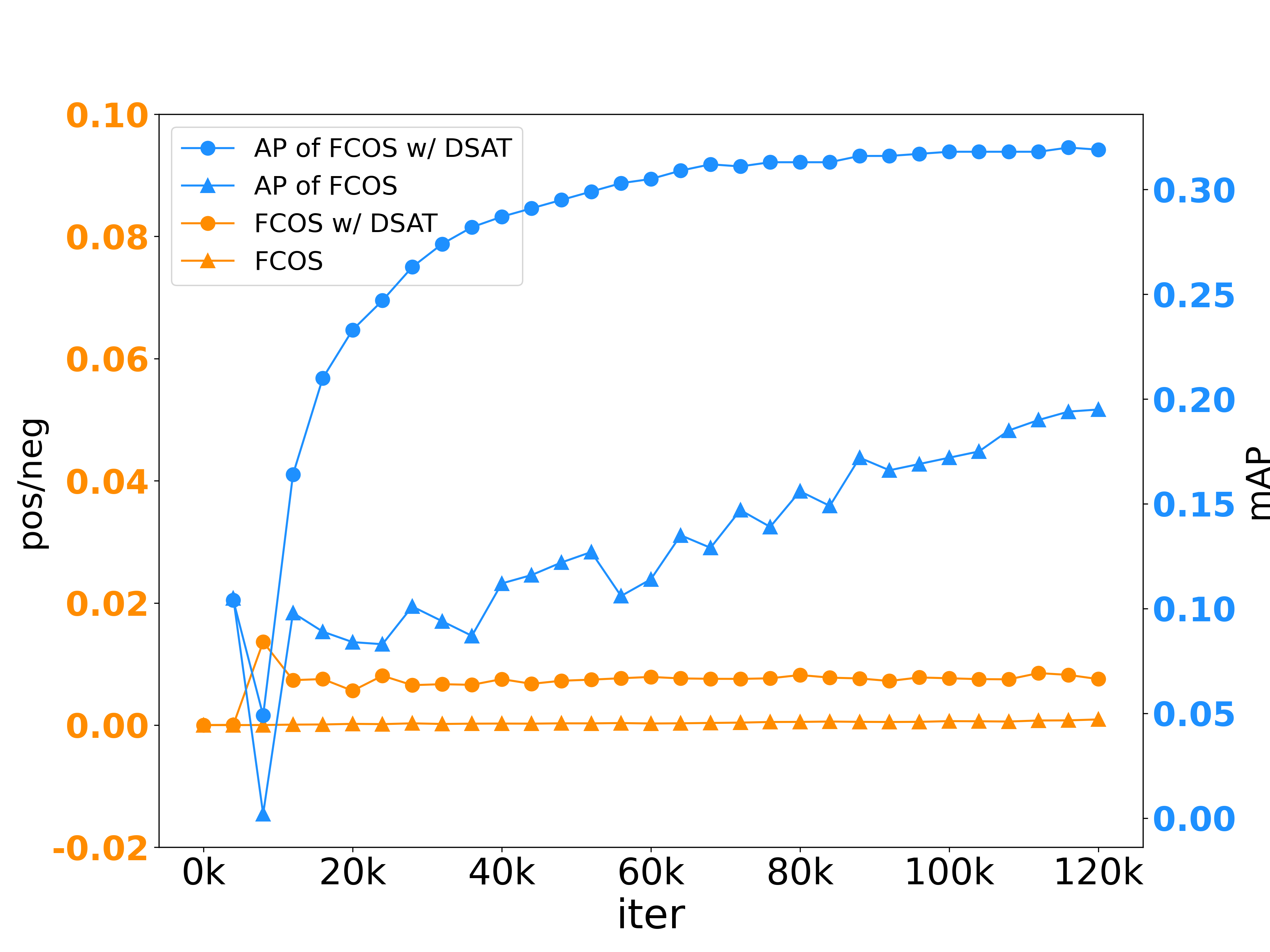}
        \label{ratio_late}
    \end{minipage}
    }
    \caption{\textbf{(a)} Precision, Recall and F1-score distribution of classification on FCOS. \textbf{(b)} Sampling ratio and corresponding mAP of w/ and w/o DSAT on FCOS. Compared with original FCOS, the proposed DSAT relieves the sampling imbalance (\textcolor[RGB]{240,145,53}{orange points}) and improves the performance (\textcolor[RGB]{66,144,247}{blue points}).}
    \label{fcos_distribution}
\end{figure}
\section{Method}
Previous semi-supervised object detection methods use the empirical strategy of selecting pseudo labels according to the ablation experiments \cite{xu2021end,liu2021unbiased}. By analyzing the distribution gap between single- and two-stage detectors, we believe that the quality and quantity of pseudo labels should be considered equally when choosing the strategy in semi-supervised learning. To this end, we propose a dynamic self-adaptive threshold (DSAT) strategy, which can dynamically adjust the strategy based on model itself and achieve an optimal trade-off between quality and quantity of pseudo labels. Besides, to assess the regression quality of pseudo labels, we develop a module named NMS-UNC that could compute uncertainty of them based on NMS. Fig.\ref{pipeline} shows the overall pipeline of the proposed semi-supervised learning for single-stage detectors (S4OD). As shown in Fig.\ref{pipeline}, we feed the dense predictions generated from teacher model into the NMS-UNC module to get the sparse pseudo labels including classification scores, boxes and uncertainty. An uncertainty threshold is used to preserve boxes with lower uncertainty. These preserve boxes are regression targets for unlabeled data. Another path use the DSAT to select boxes with higher classification scores. These boxes are assigned with one-hot labels for classification of unlabeled data.    
\subsection{Dynamic Self-Adaptive Threshold Strategy}
\label{method_2}
Since the selection of pseudo labels is of vital importance for semi-supervised learning, we propose a dynamic self-adaptive threshold strategy which can adjust the threshold automatically based on teacher model itself during the whole training process. To ensure the quality and quantity of pseudo labels, the strategy should consider both precision and recall of pseudo labels. However, precision and recall are in conflict, it's not possible to gain a higher precision with a higher recall. To reasonably make a trade-off on precision and recall, we heuristically choose the F1-score as the index to design the threshold strategy, because the F1-score denotes the harmonic mean of precision and recall. F1-score can be formulated by Eq. (\ref{f1}):
\begin{equation}
    {\rm F1-score} = \frac{2 \times Pre \times Rec}{Pre + Rec}
    \label{f1}
\end{equation}
$Pre$ and $Rec$ denote the precision and recall of the prediction. For an object detection model, we first pick a set of confidence scores (range from 0.05 to 0.95 at step 0.05). For each confidence score, we preserve all pseudo labels whose classification scores are larger than it. Then the corresponding precision, recall and F1-score on the validation set can be computed with a pre-defined IoU threshold (0.5 is used in this paper). Finally, as shown in Fig.\ref{epoch22}, the curve of precision, recall and F1-score can be obtained by connecting the corresponding points. 

To choose a proper pseudo label selecting strategy which can take both quality and quantity of pseudo labels into consideration, it should be set at the peak of the F1-score curve to achieve a relatively high quality without losing too many pseudo labels for student model. Compared with choosing a strict strategy to ensure the precision of pseudo labels, picking confidence score with respect to the F1-score peak could preserve more objects so that the imbalance of positive and negative samples can be alleviated (as shown in Fig.\ref{ratio_late}). Using the F1-score curve as the reference to choose strategy only depends on model performance, which is self-adaptive. Meanwhile, it eases the implementation greatly without the need of designing a over-complicated strategy.  

\begin{figure}[tb]
    \centering
    \subfigure[]{
    \begin{minipage}[t]{0.45\linewidth}
        \centering
        \includegraphics[width=\linewidth]{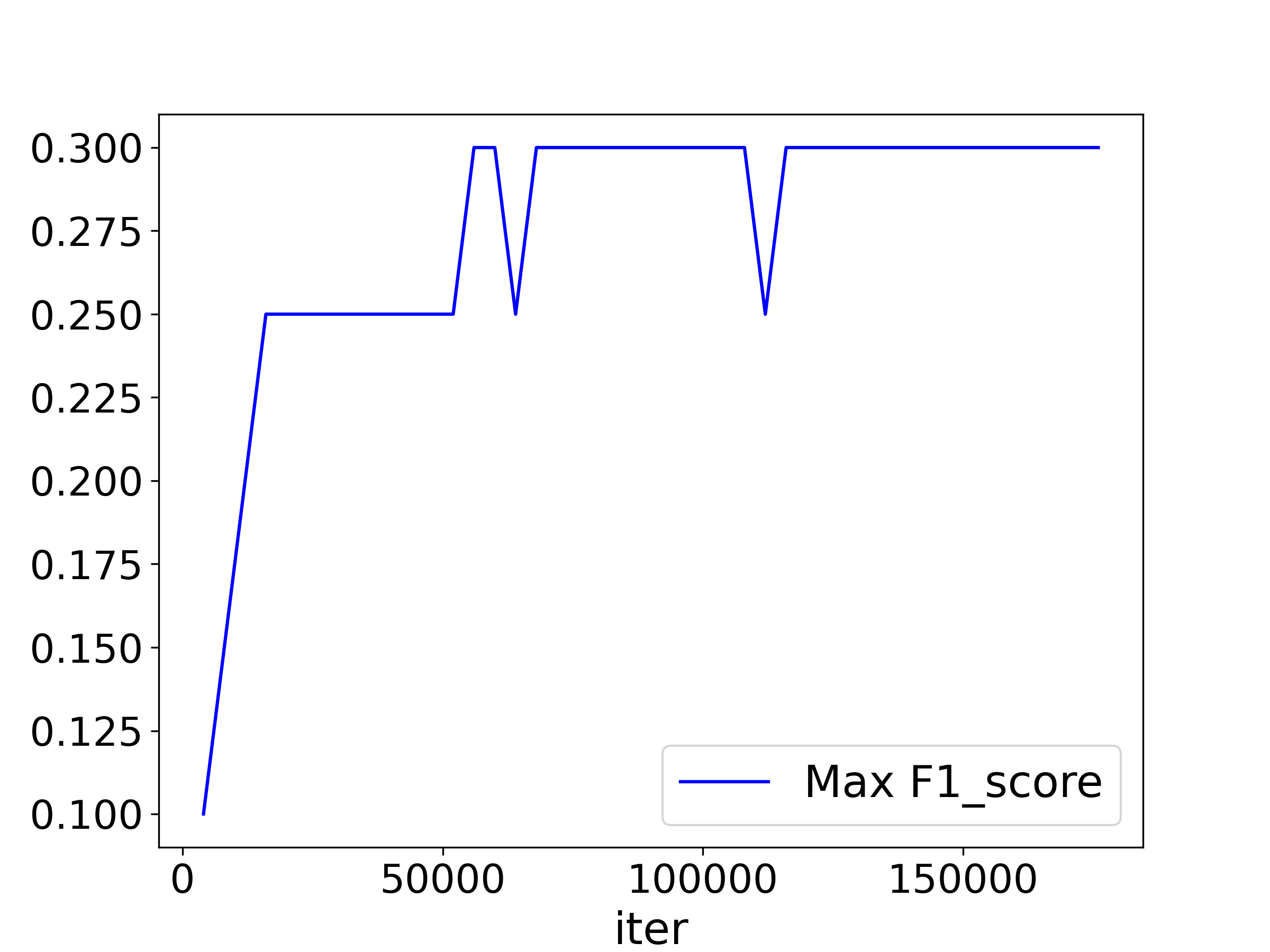}
        \label{trainging_process}
    \end{minipage}
    }
    \subfigure[]{
    \begin{minipage}[t]{0.45\linewidth}
        \centering
        \includegraphics[width=\linewidth]{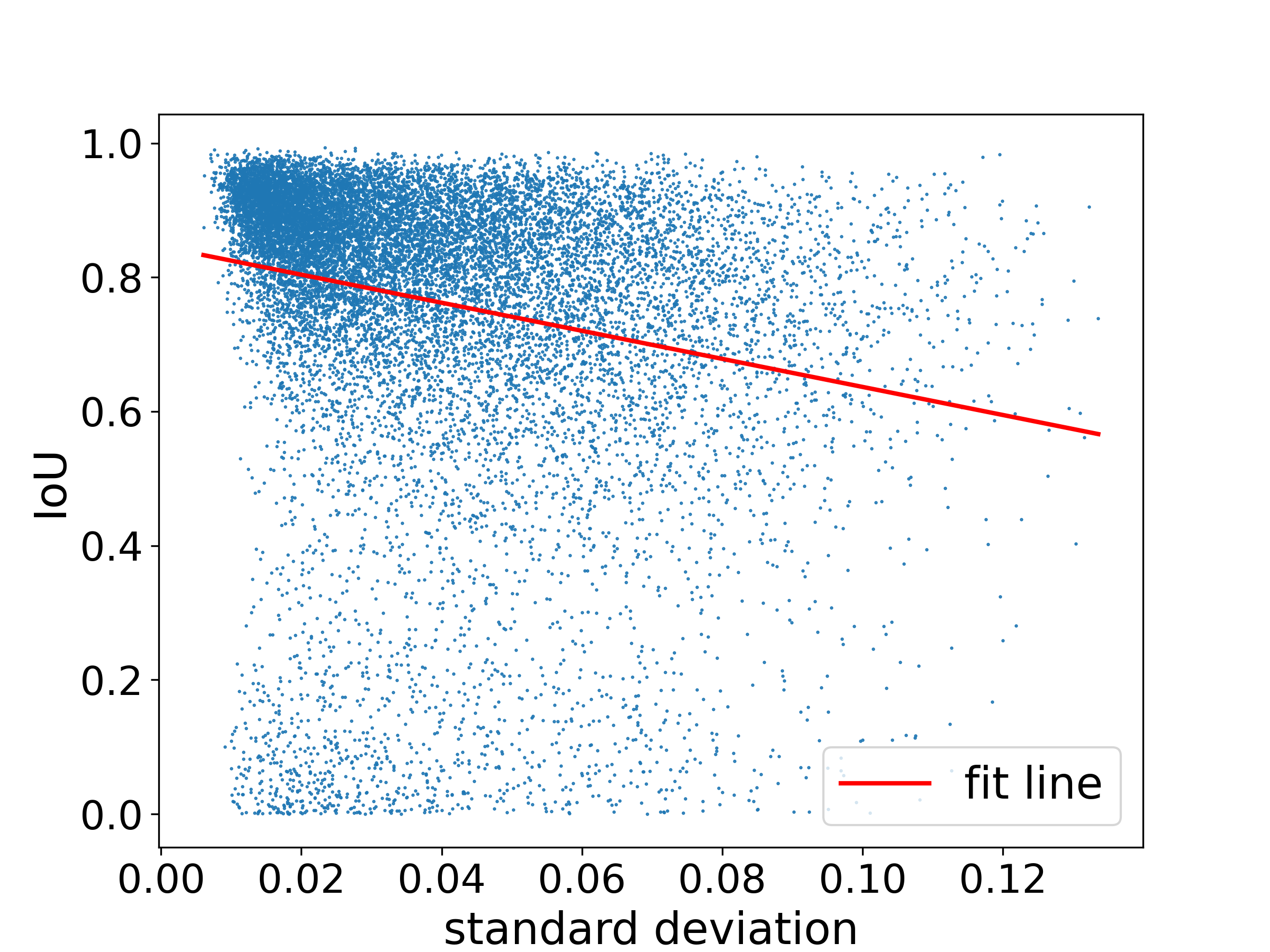}
        \label{unc_iou}
    \end{minipage}
    }
    \caption{\textbf{(a)} The confidence score corresponding to the F1-score peak during training. \textbf{(b)} The distribution between the standard deviation of redundant boxes and IoU. The fit line (in \textcolor[rgb]{1,0,0}{red}) shows the relation between them.}
    \label{training_and_unciou}
\end{figure}

As the model iterates, the performance of the model is increased accordingly. In this case, the confidence corresponding to the F1-score peak is dynamic instead of immutable during the training process (as shown in Fig.\ref{trainging_process}). As a result, using the fixed strategy would ignore more objects at the early stage and slow down the converging speed, which can aggravate the effect on the model's generalization ability. To this end, we further design a dynamic self-adaptive threshold strategy. Specifically, the F1-score curve of the teacher model can be obtained at the validation stage during training. Then we choose the confidence score corresponding to the maximum F1-score value as the filtering threshold of pseudo labels for the next training stage. Through the dynamic self-adaptive threshold strategy, the semi-supervised framework could build the most efficient information flow from the teacher model to the student model, which not only speeds up the training process but also promotes the performance.

\begin{algorithm}[tb]
\caption{NMS-UNC}
\label{nms_unc}
\LinesNumbered 
\KwIn{Dense prediction $(B_{box}, B_{score})$, IoU threshold $\delta$, Score threshold $\mu$}
\KwOut{Reserved prediction $(P_{box}, P_{score}, P_{unc})$}
$P_{box} \xleftarrow{} \varnothing$, $P_{score} \xleftarrow{} \varnothing$, $P_{unc} \xleftarrow{} \varnothing$

$B_{score} \xleftarrow{} B_{score}[B_{score}\geq \mu], B_{box} \xleftarrow{} B_{box}[B_{score}\geq \mu]$

\While{$B_{box}$ is not empty}{
    $I$ $\xleftarrow{}$ $Max$($B_{score}$)
    
    $B_{m}$ $\xleftarrow{}$  $B_{box}$[$I$], $S_{m} \xleftarrow{} B_{score}[I]$
    
    $B_{c} \xleftarrow{} \varnothing$
    
    \For{$n=0$ to $Size(B_{box})$}{
        $iou$ = $IoU(B_{box}[n], B_{m})$
    
        \If{iou $\geq$ $\delta$}{
            $B_{c} \xleftarrow{} B_{c} \cup B_{box}[n]$
        
            \textbf{del} $B_{box}[n]$
        }
    }
    \If{$Size(B_{c})$ $> 1$ }
    {
        $unc = std(B_{c})$
    
        $P_{box} \xleftarrow{} P_{box} \cup B_{m}$, $P_{score} \xleftarrow{} P_{score} \cup S_m$, $P_{unc} \xleftarrow{} P_{unc} \cup unc$
    }
}
\end{algorithm}

\subsection{NMS-UNC}
\label{method_4}
Another factor that affects the quality of pseudo-labels is regression. Different from classification, it is hard to assess the regression quality of pseudo labels in a continual domain. Inspired by the discretized continuous regression methods \cite{NEURIPS2020_f0bda020,zheng2022LD}, we propose NMS-UNC, an assessment method of box regression based on NMS (non-maximum suppression) in single-stage detectors.%

For the dense predictions of the single-stage object detectors, we find that the predicted dense bounding boxes are relatively compact for the simple objects (whose boundaries are clear) while scattered for the hard objects. Based on this observation, we believe that the standard deviation of these redundant bounding boxes that belong to one object can reflect the regression 
uncertainty of this object. As the statistics shown in Fig.\ref{unc_iou}, the bounding box with a larger standard deviation generally has a lower IoU with the ground-truth, and the fit line can indicate the relation between the proposed standard deviation with the IoU. Therefore, the standard deviation could be used to assess the regression quality. As a result, we design the NMS-UNC module which can compute the uncertainty (standard deviation) of the reserved bounding boxes during the NMS. Algorithm.\ref{nms_unc} elaborates the detail process. Consistent with the original NMS, NMS-UNC first filters out boxes whose classification scores below the score threshold $\mu$. Then, for the maximum score bounding box $B_m$ in each iteration, NMS-UNC gathers the boxes $B_c$ whose IoU with the maximum score box $B_m$ are larger than the IoU threshold $\delta$. These boxes are regarded as the redundant boxes in the original NMS module and should be removed. Before these boxes are removed, we compute the standard deviation of these boxes $B_c$ to obtain the uncertainty of $B_m$. This process can be formulated by Eq. (\ref{unc_com}):

\begin{equation}
    {\rm uncertainty} = (\sum_{i=0}^1\sum_{j=0}^1 \frac{std(B_c)[i+j]}{\mathbb{I}_{(j=0)}\bar{w} + \mathbb{I}_{(j=1)}\bar{h}})/4
    \label{unc_com}
\end{equation}
where $std \in \mathbb{R}^{4}$ denotes the standard deviation of redundant boxes $B_c$. Notably, $i$ decides the top-left point or bottom-right point of the box and $j$ indicates the coordinate axis $x$ or $y$.  $\bar{w}$ and $\bar{h}$ denote the mean width and mean height of boxes. 

After the NMS-UNC post-processing module, we can get the classification score and the regression uncertainty for each sparse pseudo label. To guarantee the quality of pseudo labels, two thresholds are set to constrain the classification score and the regression result separately. By introducing the uncertainty of pseudo labels, the unsupervised loss $L_{unsup}$ is then formulated by Eq. (\ref{unsup_loss_unc}):

\begin{equation}
    L_{unsup} = \frac{1}{N_{pos}^c}\sum_{x,y}L_{cls}(p_{x,y}, c^*_{x,y}) + \frac{1}{N_{pos}^b}\sum_{i}^{N_{pos}^b}L_{reg}(t_{i}, \hat{b}_{i})
    \label{unsup_loss_unc}
\end{equation}
where $N_{pos}^c$ and $c^*$ are the number of positive samples and classification target according to the pseudo classification label whose scores are higher than $\sigma_{cls}$. $\hat{b}$ is the regression targets whose uncertainty is smaller than regression threshold $\sigma_{unc}$. $N_{pos}^b$ is the number of positive samples assigned from these regression targets $\hat{b}$.

\section{Experimental Results}
\subsection{Implementation Details}
\noindent\textbf{Dataset and Metrics.} All experiments are conducted on the MS-COCO \cite{lin2014microsoft} dataset including 80 object categories and 118k images. For semi-supervised training, we follow the setting from Xu et al. \cite{xu2021end} and split the COCO-train2017 set (all labeled) into multiple datasets with different labeled percentages: 1\%, 5\% and 10\%. Besides, we use the COCO-unlabel2017 to conduct incremental experiments including 123k images. The COCO-val2017 containing 5,000 images is used to evaluate the mean average precision (mAP) metrics.

\noindent\textbf{Training Details.} We implement the anchor-free single-stage detector (FCOS \cite{tian2019fcos}) and the anchor-based single-stage detector (RetinaNet \cite{lin2017focal}) based on MMdetection \cite{mmdetection}. Different from classification and regression branches in anchor-based detectors, the dense output from FCOS includes another centerness branch. We use the classification score rescaled by centerness as the final score and compare the score with the threshold to select high-quality pseudo labels. ResNet-50 \cite{he2016deep} with FPN \cite{lin2017feature} is used and the student model is trained by the SGD optimizer for 180K iterations with 0.9 momentum and 0.0001 weight decay. The learning rate is set to 0.01 with a warm-up strategy and decays by 0.1 at iteration 120K and 160K. For teacher model, the EMA update rate is set to 0.999. The same data-augmentation method used by Xu et al. \cite{xu2021end} is applied in our experiments. Batchsize is 32 and 8 Nvidia Tesla P40 GPUs are used for training. Unless otherwise specified, all experiments are conducted as above.

For semi-supervised baseline, we set the fixed classification threshold $\sigma_{cls}$ to 0.5 as it achieves the best performance in Table~\ref{classification_ablation}. For dynamic threshold setting, we evaluate the F1-score at each 4,000 iterations and update the selecting strategy according to the latest F1-score peak. As for uncertainty threshold, we use $\sigma_{unc}=0.08$ according to the ablation study result in Table~\ref{uncertainty_ablation}. The original score threshold $\mu$ is 0.05 in NMS, which would preserve many noise bounding boxes for uncertainty computing in NMS-UNC. Therefore, we increase the score threshold $\mu$ to 0.4 for preserving the valuable bounding boxes to assess the regression quality. For other hyper-parameters in FCOS and RetinaNet, we follow the default setting in MMdetection.

\noindent\textbf{Inference Details.} During the semi-supervised training, the teacher model always performs better than student model. As a result, we choose the teacher model as the final model  and use the same sizes of input images as training. Because only the teacher model is used to inference, the semi-supervised learning would not increase the inference time. Besides, to make a fair comparison, NMS score threshold $\mu$ is set to 0.05 in inference and evaluation. 

\begin{table}[tb]
\begin{center}
\caption{Results on different labeled percentages.}
\label{overall_result}
\setlength{\tabcolsep}{0.5mm}{
\begin{tabular}{llllll}
\toprule
Model   & Supervised &                 & 1\% & 5\% & 10\% \\
\midrule
\multirow{4}{*}{Faster-RCNN} & Fully & Baseline & 10.0 & 20.92 & 26.94 \\\cmidrule(lr){2-6}
&\multirow{3}{*}{Semi} & STAC\cite{sohn2020simple} & 13.97(+3.97) & 24.38(+3.46) & 28.64(+1.70)\\
&&Unbiased\cite{liu2021unbiased} &20.75(+10.75) & 28.27(+7.35) & 31.50(+4.56)\\
&&SoftTeacher\cite{xu2021end} & 20.46(+10.46) & 30.74(+9.82) & 34.04(+7.10)\\
\midrule
\multicolumn{1}{l}{\multirow{2}{*}{FCOS\cite{tian2019fcos}}} & Fully & Baseline & 13.6 & 23.5 & 28.1 \\ \cmidrule(lr){2-6}
\multicolumn{1}{l}{}                      &  Semi & \textbf{Ours}  & \textbf{23.7}(+10.1) & \textbf{32.3}(+8.8) & \textbf{35.0}(+6.9) \\
\midrule
\multirow{2}{*}{RetinaNet\cite{lin2017focal}}                & Fully & Baseline & 12.2 & 22.7 & 27.4 \\ \cmidrule(lr){2-6}
 & Semi & \textbf{Ours}  & \textbf{20.1}(+7.9)  & \textbf{30.0}(+7.3)  &  \textbf{32.9}(+5.5) \\
\bottomrule
\end{tabular}}
\end{center}
\end{table}

\begin{table}[tb]
\begin{center}
\caption{Experiments of using all labeled data of COCO-train2017 and the extra unlabeled data of COCO-unlabel2017. }
\label{additional}
\setlength{\tabcolsep}{3mm}{
\begin{tabular}{llll}
\toprule
            & Labeled                         & Unlabeled                         & mAP\\
\midrule
Baseline    &     COCO-train2017          &           --                    &  37.9 \\
Ours & \multirow{1}{*}{COCO-train2017} & \multirow{1}{*}{COCO-unlabel2017} &   \textbf{42.8}(+4.9) \\
\bottomrule
\end{tabular}}
\end{center}
\end{table}

\subsection{Results Comparison}

To verify the effectiveness of semi-supervised learning in single-stage detectors, we conduct elaborate experiments and report the results in Table~\ref{overall_result} and Table~\ref{additional}. Table~\ref{overall_result} shows that the proposed semi-supervised learning has achieved a significant improvement compared with fully-supervised learning. The semi-supervised setting outperforms the fully-supervised training by more than +5.5\% AP, +10.1\% at most. These improvements convince that by automatically adjusting the selective strategy  and using the proposed NMS-UNC to choose regression targets, single-stage object detectors can achieve excellent improvements as well.

Specifically, for anchor-free object detector (FCOS), the proposed method consistently boosts the performances by +10.1\%, +8.8\% and +6.9\% AP in 1\%, 5\% and 10\% labeled data setting correspondingly. For anchor-based single-stage detector (RetinaNet), the improvements are +7.9\%, +7.3\% and +5.5\% on AP. 

In addition, we conduct incremental experiments on all labeled data of COCO-train2017 set on FCOS. As shown in Table~\ref{additional}, when using all data of COCO-train2017 set, the fully-supervised learning achieves 37.9\% AP. By leveraging the unlabeled data of COCO-unlabel2017 additionally, our S4OD method reaches 42.8\% on AP, which is 4.9\% higher than the baseline. 
\begin{table}[tb]
\begin{center}
\caption{Ablation study on the effectiveness of the proposed modules.}
\label{FCOS_result}
\setlength{\tabcolsep}{3mm}{
\begin{tabular}{lccccc}
\toprule
Method                           & DSAT & NMS-UNC & AP  & AP$_{.5}$  & AP$_{.75}$ \\
\midrule
Fully-supervised                       &                   &         & 23.5 & 39.5 & 24.0 \\ 
\midrule
\multirow{4}{*}{Semi-supervised} &                   &         & 29.8 & 47.9 & 31.3 \\ 
                                 &                   &\checkmark& 31.3 & 49.1 & 33.0 \\ 
                                 &    \checkmark     &         & 31.6 & 49.2 & 33.3 \\ 
                                 &    \checkmark     &\checkmark& \textbf{32.3} & \textbf{50.1} & \textbf{34.0} \\ \bottomrule

\end{tabular}}
\end{center}
\end{table}

\begin{table}[tb]
\begin{center}
\caption{Ablation study on the fixed classification threshold $\sigma_{cls}$.}
\label{classification_ablation}
\setlength{\tabcolsep}{2mm}{
\begin{tabular}{clll}
\toprule
$\sigma_{cls}$ & AP & AP$_{.5}$ & AP$_{.75}$\\\midrule
0.4       &  29.3 & 47.2 & 30.6\\
0.5       &  \textbf{29.8} & \textbf{47.9} & \textbf{31.3}\\
0.6       &  29.1 & 46.2 & 30.5\\
0.7       &  24.2 & 39.0 & 25.2\\
\midrule
DSAT (ours)&  \textbf{31.6}(+1.8) & \textbf{49.2}(+1.3) & \textbf{33.3}(+2.0)\\
\bottomrule
\end{tabular}}
\end{center}
\end{table}

\subsection{Ablation Studies}

\noindent\textbf{Proposed Modules.} We design ablation experiments on FCOS to analyze the effectiveness of the proposed DSAT strategy and NMS-UNC module. Table~\ref{FCOS_result} shows that both the DSAT strategy and the NMS-UNC module outperform the semi-supervised method in all different labeled percentages. Most importantly, the model achieves the best results when combining the two modules, which means that the proposed modules are orthogonal.

\noindent\textbf{Fixed Classification Threshold.}
For the fixed classification threshold, we conduct different classification threshold settings on FCOS to find the best-performed classification threshold. As shown in Table~\ref{classification_ablation}, when the classification threshold is 0.5, the models achieve the best performance. Notably, compared with the lower thresholds, the models are more sensitive to the higher thresholds. When increasing the threshold to 0.7, the performance drops sharply. These results support the view of fewer pseudo labels lead to a more severe sampling imbalance of positive and negative. It would waste expensive time and computing cost to find a suitable classification threshold by ablation experiments, which inspires us to design a dynamic self-adaptive threshold strategy. The bottom row lists our DSAT results. Compared with the best fixed threshold setting, the proposed DSAT achieves +1.8\% improvement on AP.

\noindent\textbf{Uncertainty Threshold.} By introducing the proposed NMS-UNC module, we can calculate the regression uncertainty for each pseudo label. A well-chosen uncertainty threshold can meet the demand of filtering out the pseudo labels with higher uncertainty. Table~\ref{uncertainty_ablation} shows that when the threshold is 0.08, the model achieves the best performance. The similar conclusion can be drawn from Fig.\ref{unc_iou}, the scatters with higher IoU become more sparser when the standard deviation (uncertainty) is higher than 0.08. For uncertainty threshold 0.05, it would abandon a large number of well-regressed bounding boxes and affect the generalization of models. As a result, we use $\sigma_{unc}=0.08$ in our experiments for optimal performances.
\begin{table}[tb]
\begin{center}
\caption{Ablation study on the uncertainty threshold $\sigma_{unc}$.}
\label{uncertainty_ablation}
\setlength{\tabcolsep}{3mm}{
\begin{tabular}{llll}
\toprule
$\sigma_{unc}$ & AP & AP$_{0.5}$ & AP$_{0.75}$\\\midrule
0.05       &  31.0 & 48.3 & 32.9 \\
0.08       &  \textbf{31.3} & \textbf{49.1} & \textbf{33.0} \\
0.10       &  31.0 & 48.8 & 32.9 \\
\bottomrule
\end{tabular}}
\end{center}
\end{table}

\noindent\textbf{Score Threshold in NMS-UNC.} In the original NMS, a score threshold $\mu$ is used to select high quality boxes and reduce the computation. To ensure the recall of detectors, $\mu$ usually sets to 0.05. In this paper, to assess the regression quality for selecting pseudo labels, we proposed the NMS-UNC module based on NMS, which computes the standard deviation of these redundant boxes as its uncertainty before removing them. To avoid the affect caused by the low-quality boxes in computing uncertainty, the score threshold $\mu$ should be increased properly. As a result, we design ablation experiments on the score threshold $\mu$ and list the result in Table~\ref{score_ablation}. From Table~\ref{score_ablation}, the NMS-UNC module achieves the best performance when the score threshold $\mu$ is 0.4. Besides, $\mu=0.4$ is more efficient in computing compared with $\mu=0.3$, because $\mu=0.3$ would involve more boxes.

\begin{table}[t]
\begin{center}
\caption{Ablation study on the score threshold $\mu$ in NMS-UNC.}
\label{score_ablation}
\setlength{\tabcolsep}{3mm}{
\begin{tabular}{llll}
\toprule
$\mu$ & AP & AP$_{0.5}$ & AP$_{0.75}$\\\midrule
0.3       & 31.0 & 48.4 & 33.0 \\
0.4       & \textbf{31.3} & \textbf{49.1} & \textbf{33.0} \\
0.5       & 30.7 & 48.0 & 32.2 \\
\bottomrule
\end{tabular}}
\end{center}
\end{table}

\begin{table}[tb]
\begin{center}
\caption{Running time analyze of the proposed module.}
\label{runtime}
\setlength{\tabcolsep}{3mm}{
\begin{tabular}{ccll}
\toprule
DSAT & NMS-UNC & sec/iter & total time(sec) \\
\midrule
                  &         & 2.470    & 444600           \\
       \checkmark &         & 2.520    & 453600(+2.02\%)  \\
       \checkmark &   \checkmark      & 2.607    & 469260(+5.55\%)  \\
\bottomrule
\end{tabular}}
\end{center}
\end{table}
\subsection{Discussion}

\noindent\textbf{Time Comparison.} In this section, we make a time analysis on the proposed DSAT and NMS-UNC. The increased time cost of DSAT strategy comes from the F1-score computing. As shown in Table~\ref{runtime}, the time cost of F1-score computing is negligible because only 2.02\% training time increased. NMS-UNC module needs to compute the box uncertainty based on the original NMS module. We integrate the uncertainty computing process with the original CPU version NMS instead of the CUDA version NMS for simplified implementation, which would slow down the training. However, this process can be accelerated by utilizing CUDA Core if necessary. The increased time is acceptable compared with the whole training process. Moreover, the uncertainty is only computed in training, there is no additional time cost for inference and evaluation.

\noindent\textbf{Application Value.}
The proposed DSAT module is heuristic, because it has no hypothesis to the model and data, which means that the proposed strategy could help the semi-supervised learning to be applied in most engineering object detection tasks. Besides, although the proper filter strategy can be determined by conducting sophisticated ablation experiments, the performance of the fixed classification threshold is still lower than that of the DSAT strategy (as shown in the second and fourth rows of Table~\ref{FCOS_result}). In conclusion, our DSAT strategy outperforms the fixed threshold strategy prominently not only in the implementation cost but also in performance.

\section{Conclusions}
In this work, we point out that the improper strategy of selecting pseudo labels would intensify the imbalance of positive and negative samples of single-stage detectors in semi-supervised learning. As a result, how to choose a proper strategy is critical to semi-supervised learning for single-stage object detectors. To guarantee both the quality and quantity of pseudo labels, we propose a dynamic self-adaptive threshold strategy, which could adjust the selecting strategy according to the model performance automatically. Besides, we also design a measurement method of box regression quality for single-stage detectors. Extensive experiments in the MS-COCO dataset convince that the proposed method achieves state-of-the-art performance in semi-supervised single-stage object detection.   

\clearpage
%
%
\bibliographystyle{splncs04}
\bibliography{submission}
\end{document}